\documentclass[conference,a4paper]{IEEEtran}
\IEEEoverridecommandlockouts

\usepackage[hidelinks]{hyperref}
\usepackage[cmex10]{amsmath}
\usepackage{amssymb,amsfonts}
\usepackage{dblfloatfix}

\usepackage[ruled,vlined]{algorithm2e}
\usepackage{graphicx}
\graphicspath{{Figures/PDF/}{Figures/PNG/}}
\usepackage{flushend}
\usepackage{booktabs}
\usepackage{siunitx}
\usepackage[numbers,compress]{natbib}
\usepackage{texnames}
\usepackage{bm,bbm}
\usepackage{orcidlink}

\begin{document}


\title{Patch-GAN Transfer Learning with Reconstructive Models for Cloud Removal}

\author{	\IEEEauthorblockN{Wanli Ma\orcidlink{0000-0003-4088-9193}, Oktay Karakuş\orcidlink{0000-0001-8009-9319}, Paul L. Rosin\orcidlink{0000-0002-4965-3884} }
	\IEEEauthorblockA{\textit{School of Computer Science and Informatics}\\Cardiff University, Cardiff, CF24 4AG, UK. \\
		\{maw13, karakuso, rosinpl\}@cardiff.ac.uk}
}

\maketitle
\begin{abstract}
	Cloud removal plays a crucial role in enhancing remote sensing image analysis, yet accurately reconstructing cloud-obscured regions remains a significant challenge. Recent advancements in generative models have made the generation of realistic images increasingly accessible, offering new opportunities for this task. Given the conceptual alignment between image generation and cloud removal tasks, generative models present a promising approach for addressing cloud removal in remote sensing. In this work, we propose a deep transfer learning approach built on a generative adversarial network (GAN) framework to explore the potential of the novel masked autoencoder (MAE) image reconstruction model in cloud removal. Due to the complexity of remote sensing imagery, we further propose using a patch-wise discriminator to determine whether each patch of the image is real or not. The proposed reconstructive transfer learning approach demonstrates significant improvements in cloud removal performance compared to other GAN-based methods. Additionally, whilst direct comparisons with some of the state-of-the-art cloud removal techniques are limited due to unclear details regarding their train/test data splits, the proposed model achieves competitive results based on available benchmarks.
\end{abstract}


\section{Introduction}
Cloud removal is an essential topic in remote sensing and earth observation, as clouds frequently obscure the Earth's surface in optical satellite images, hindering the effective analysis of remote sensing data. For example, images from satellite platforms such as Landsat, Sentinel-2, and MODIS are highly susceptible to cloud interference. Accurate cloud removal methods are crucial for obtaining reliable data used in environmental monitoring, land-use analysis \cite{10015046, shen2019spatiotemporal}, agriculture \cite{ sare2024joint}, and other applications \cite{shen2014effective}. Therefore, it is vital to develop efficient cloud removal methods for optical remote sensing images.

With the rapid development of generative and reconstructive models, these models, which exhibit exceptional capabilities, are now capable of creating images that are virtually indistinguishable from real ones. The Masked Autoencoder (MAE) \cite{he2022masked} is one such model, designed to reconstruct the original image from its partial observations. The obstruction caused by any obstacle (such as cloud cover in remote sensing images) restricts the observable information to only partial details. MAE effectively reconstructs the hidden information, making it particularly suitable for tasks like cloud removal, where the goal is to eliminate the clouds and generate the content of the corresponding area. In this sense, cloud removal is akin to image reconstruction, but it presents additional challenges, as the input cloudy image may not provide sufficient information for accurate reconstruction. Considering that the MAE model has the capability to reconstruct an entire image using only partial image information as input, the motivation of this paper is to fine-tune a MAE-based pre-trained reconstructive model (originally trained on a large natural image dataset) and transfer its knowledge to the cloud removal task. This approach aims to extract information from cloudy multi-spectral remote sensing images and generates cloud-free counterparts, ultimately supporting remote sensing image analysis. 

Remote sensing images are more complex than natural images due to their inherent diversity and variability. Unlike natural images, which often depict a limited number of static objects, remote sensing images typically contain multiple objects—such as buildings, vegetation, roads, and bodies of water—that vary in location and appearance. These objects are not fixed, and their spatial arrangement can change significantly between images. This variability makes it challenging to apply traditional image generation models directly. To address this complexity, we opted for a patch-based GAN structure, rather than an image-wise discriminator. In a patch-based GAN, the discriminator evaluates whether each individual patch (a small, localized portion of the image) is real or fake, rather than assessing the entire image at once. This approach allows the model to focus on local details and effectively handle the diverse structures and spatial relationships present in remote sensing imagery. By processing smaller patches independently, the model can better capture the intricate features of different objects and their varying locations, leading to improved accuracy in reconstructing missing or obscured information.

Specifically, the contributions of this paper are as follows:  
(1) We propose a novel deep transfer learning approach integrated with a patch-based GAN framework, which effectively handles the complexity of remote sensing images. This approach allows for localized evaluation of image patches, improving the accuracy of cloud removal by capturing the intricate features and spatial variability typical of remote sensing data. 
(2) We identify the inherent similarity between the masked autoencoder-based reconstruction task and cloud removal. By recognizing this connection, we successfully explore and transfer the knowledge of reconstructive models, pre-trained on large natural image datasets, to the specific challenge of cloud removal in remote sensing imagery. This transfer not only enhances the cloud removal process but also leverages the power of pre-trained models to address the unique challenges posed by remote sensing data.

\section{Related Work}

Clouds in multi-spectral remote sensing imagery can be classified into two categories: \textit{thick} and \textit{thin} clouds. When an area is covered by thick clouds, the information available for that region in optical remote sensing images becomes very limited. Widely used approaches for the thick cloud removal task are to use multi-modal or multi-temporal remote sensing data to collect additional information \cite{10015046, meraner2020cloud, chen2019blind, chen2020thick}. On the other hand, image enhancement and filter techniques are used for thin cloud removal \cite{liu2014thin, shen2014effective}.  Before applying deep learning in cloud removal, traditional image processing techniques addressed the task through pixel correction. Gafurov \textit{et al.} \cite{gafurov2009cloud} proposed a method involving six successive steps to estimate pixel coverage by utilizing various temporal and spatial information sources. Wang \textit{et al.} \cite{5677195} proposes using grayscale morphological operations and contrast expansion based on traditional homomorphic filtering to preserve surface information, enhance image clarity, and remove cloud effects.

With the rapid development of deep learning, the application of neural networks in cloud removal tasks has become more widespread where GANs are commonly used for this purpose in the last decade \cite{singh2018cloud, li2020thin}. GANs, in particular, comprise two components — a \textit{generator} and a \textit{discriminator} — that are trained concurrently through a competitive process. The generator is tasked with producing cloud-free images, while the discriminator differentiates between real data (from the original dataset) and synthetic data (created by the generator). 

Masked Autoencoder (MAE) \cite{he2022masked} is an autoencoder (encoder-decoder) technique that reconstructs the original image based on its partial observation. MAE's encoder is a standard vision transformer (ViT) \cite{dosovitskiy2020image} but it operates solely on visible, unmasked patches. Similar to a standard ViT, the encoder embeds patches through a linear projection combined with positional embeddings and processes the resulting set using a series of Transformer blocks. During the MAE training phase, masked patches are excluded, meaning only a small subset of the full set is utilized. This approach enables the training of very large encoders while requiring only a fraction of the computational and memory resources.
The MAE's decoder is also a ViT, designed to process the full set of tokens, which includes both the encoded visible patches and the mask tokens. The decoder generates a reconstructed image, filling in the areas that were masked in the input image. Although the MAE was primarily designed for image representation, with the encoder typically used for downstream tasks, its reconstruction capability is also highly effective, particularly when combined with GAN loss.

Deep transfer learning is a machine learning technique that utilizes the knowledge learned from one task to improve the performance of another related task \cite{tan2018survey}. 
Transfer learning has been utilized in the literature for tasks such as cloud detection \cite{pang2023convolutional} and cloud removal \cite{ahn2020cloud}, but its application to cloud removal remains limited and is still in its early stages.

\section{Methodology}
\begin{figure*}
    \centering
    \includegraphics[width=\linewidth]{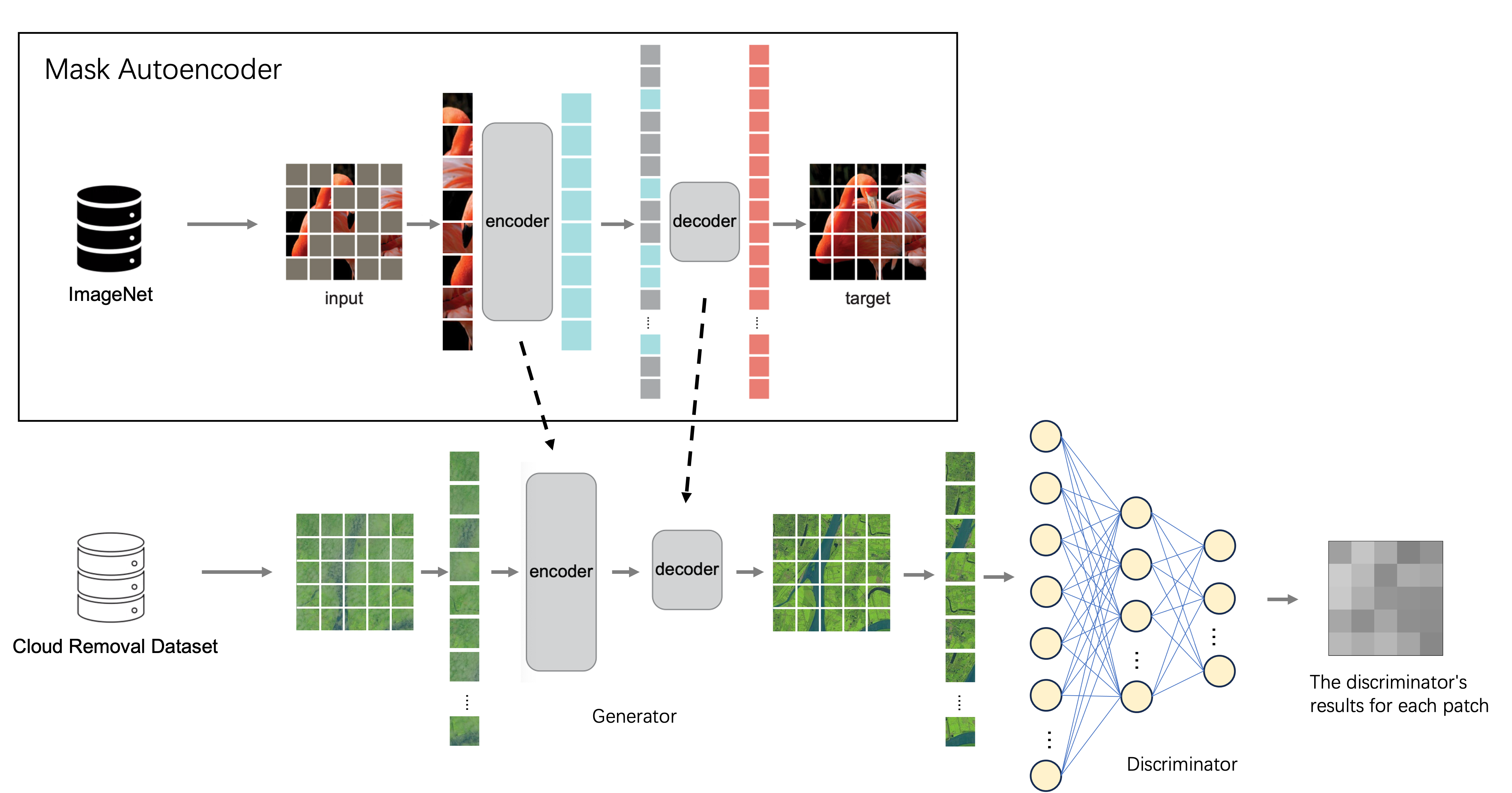}
    \caption{The overall framework of the proposed deep transfer learning. }
    \label{fig:overall_diagram}\vspace{-0.24cm}
\end{figure*}
The proposed deep transfer learning approach uses a GAN architecture, which comprises both a generator and a discriminator. Figure \ref{fig:overall_diagram} illustrates the overall structure of the proposed method.

\subsection{Generator}
The generator is a cloud removal model featuring a vision transformer (ViT-large) encoder and a ViT decoder. It is initially pre-trained for an image reconstruction task using MAE on the ImageNet dataset. As the model is built on a vision transformer, the input image is randomly cropped to 224×224 during the training phase to match the required input format. The cropped image is then divided into 196 patches, each measuring 16×16 pixels, before putting it into the encoder. The decoder generates predicted cloud-free patches based on the input patches, which are then combined to form a complete image. 

\subsection{Discriminator}
The Discriminator network is a fully connected neural network as shown in Figure \ref{fig:overall_diagram}. The generated cloud-free image is partitioned into 196 patches, each measuring 16×16 pixels. Each patch has RGB channels, which are flattened into a one-dimensional signal with a length of 768. The output of the discriminator is a one-dimensional signal of length 196, where each value represents whether the corresponding patch is predicted to be real or fake.

\subsection{Loss}
The generator is guided by both MSE and GAN losses. The ground truth is used to supervise the generator by minimizing the MSE loss $\mathcal{L}_{MSE}$ between the predictions $p$ and the ground truth $g$. 
\begin{equation}
\label{equ:equ1}
\mathcal{L}_{MSE} = \frac{1}{C \times W \times H} \sum_{k=1}^{C \times W \times H} |p_k - g_ks|^2 ,
\end{equation}
where $C$, $W$ and $H$ refer to the channel, weight and height of features, respectively. 

Another component of the losses is the GAN loss $L_{GAN}$, which consists of the discriminator loss $L_{GAN}^{D}$ and generator loss $L_{GAN}^{G}$. Thus GAN loss is defined as
\begin{equation}
    L_{GAN} = L_{GAN}^{D} + L_{GAN}^{G},
\end{equation}
where the discriminator loss is
\begin{equation}
    L_{GAN}^{D} = -\frac{1}{P} \sum_{k=1}^{P}[\log x_k + \log \left(1 - \hat{x_k}\right)],
\end{equation}
with $P$ represents the number of patches in the image. $x$ represents the Discriminator's prediction for real input, and $\hat{x}$ represents its prediction for generated input.
The generator aims to deceive the discriminator, and the GAN loss for the generator is expressed as follows:
\begin{equation}
    L_{GAN}^{G} = -\frac{1}{P} \sum_{k=1}^{P}\log (\hat{x}_k) 
\end{equation}
where $P$ represents the number of patches in the image and $\hat{x}$ represents its prediction for generated input.

We used a layer-wise learning rate decay strategy in the fine-tuning where different layers of the neural network are assigned different learning rates during the training process. Earlier layers (closer to the input) are given smaller learning rates, while later layers (closer to the output) receive higher learning rates. This is because the weights in the earlier layers usually contain more general features (such as edges, textures, etc.) learned from the pre-training stage, which may not need to be adjusted as much. In contrast, the later layers, which tend to capture more task-specific features, may require more significant updates to adapt to the fine-tuning task.

\begin{figure}[htbp]
    \centering
    \includegraphics[width=0.23\linewidth]{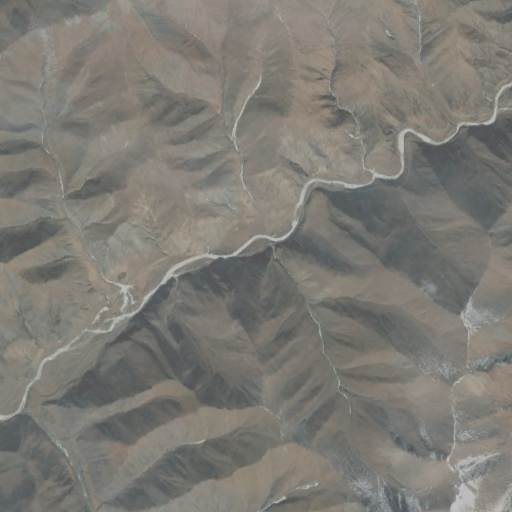}
    \includegraphics[width=0.23\linewidth]{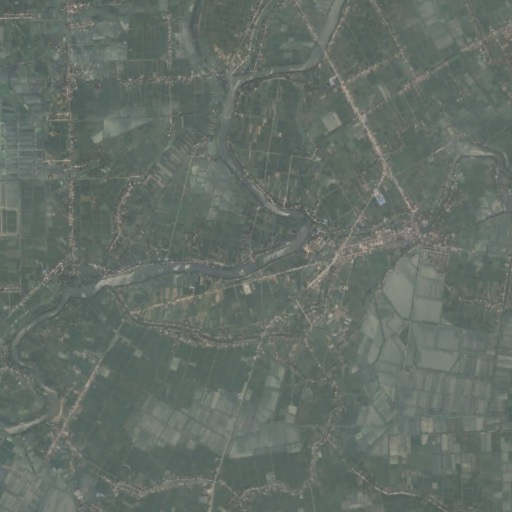}
    \includegraphics[width=0.23\linewidth]{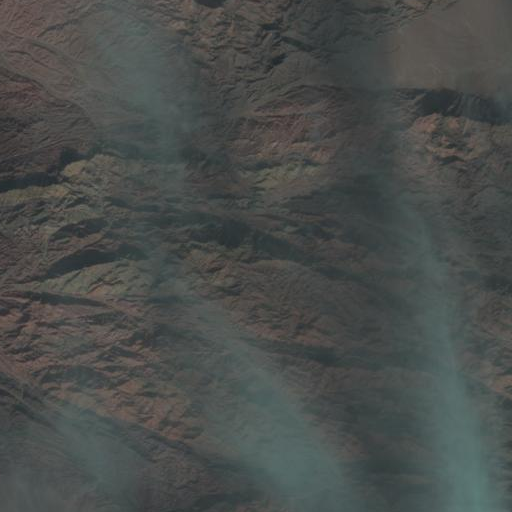}
    \includegraphics[width=0.23\linewidth]{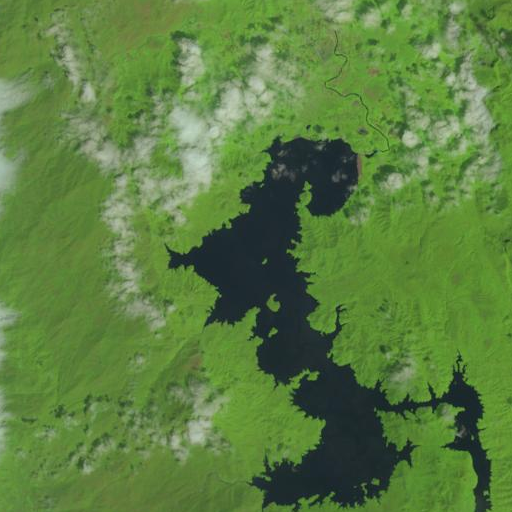}
    
    \vspace{3pt}
    \includegraphics[width=0.23\linewidth]{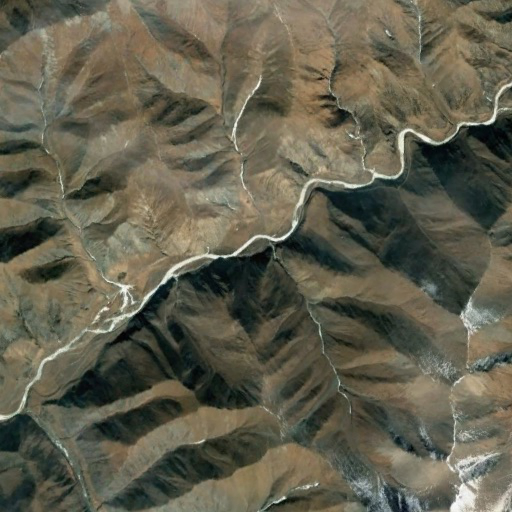}
    \includegraphics[width=0.23\linewidth]{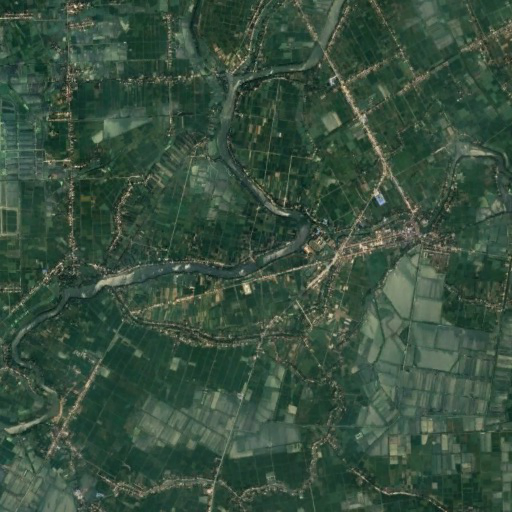}
    \includegraphics[width=0.23\linewidth]{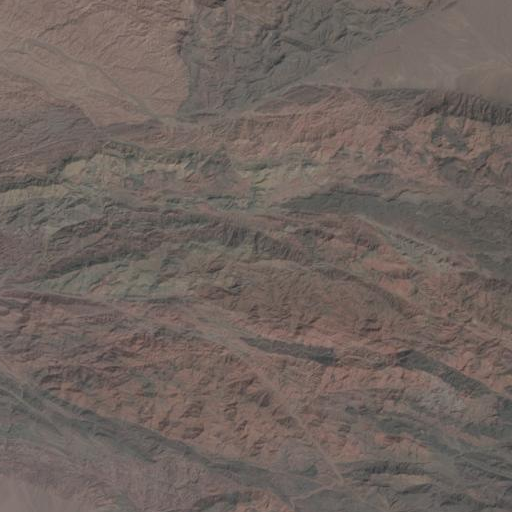}
    \includegraphics[width=0.23\linewidth]{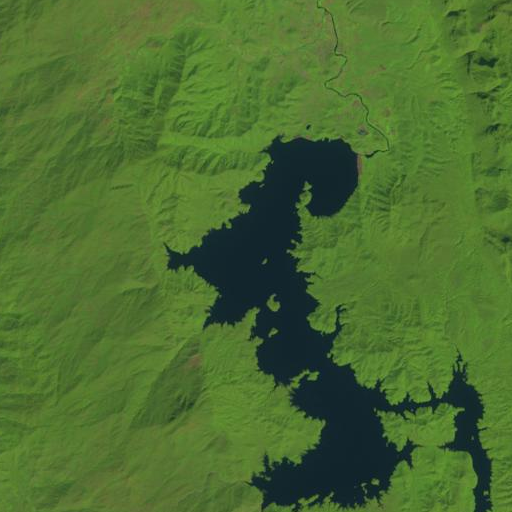}
    \caption{Dataset samples of RICE dataset. The top row displays cloudy images, while the bottom row shows cloud-free images. The first two samples on the left are from RICE1, while the two samples on the right are from RICE2. }
    \label{fig:rice_dataset}\vspace{-0.25cm}
\end{figure}

\section{Experiments and Results}
\subsection{Dataset}
We used the RICE dataset \cite{lin2019remote} to evaluate the proposed method. It is divided into two subsets: RICE1 comprises 500 image pairs, each consisting of a cloud-covered image and its corresponding cloud-free image, both with dimensions of 512×512. A total of 400 image pairs are used for training, and 100 image pairs are reserved for testing. The dataset is collected from Google Earth. RICE2 includes 736 image sets, with each set containing three 512×512 images: a reference cloud-free image, a cloud-covered image, and the corresponding cloud mask. A total of 588 image pairs are used for training, and 148 image pairs are reserved for testing. RICE1 and RICE2 are obtained from Google Earth and Landsat 8OLI/TIRS data, respectively. Following \cite{pan2020cloud}, we used 1/5 training data for validation during the training phase. Examples of data from the RICE dataset are displayed in Figure \ref{fig:rice_dataset}.

\subsection{Experimental Setup}
Our experiments were conducted using PyTorch with a mini-batch ADAM optimizer. The base model\footnote{Model link: \url{https://dl.fbaipublicfiles.com/mae/visualize/mae_visualize_vit_large_ganloss.pth}} we fine-tuned is the MAE-large model, enhanced with GAN loss. The experiments ran on an NVIDIA 3060 GPU, and the models were evaluated using peak signal-to-noise ratio (PSNR) and structural similarity index (SSIM) metrics. 
\begin{equation}
\mbox{PSNR}=10 \log _{10}\left(\frac{\left(MAX_I\right)^2}{M S E}\right)
\end{equation}
where $MAX_I$ is the maximum value that a pixel can take, and $MSE$ is
\begin{equation}
M S E=\frac{1}{H \times W} \sum_{i=1}^H \sum_{j=1}^W(X(i, j)-Y(i, j))^2
\end{equation}
where $X$ represents the predicted, and $Y$ the ground truth images. $\operatorname{SSIM}$ is also defined as
\begin{equation}
\operatorname{SSIM}=\frac{2 \mu_X \mu_Y+C_1}{\mu_X^2+\mu_Y^2+C_1} \cdot \frac{2 \sigma_X \sigma_Y+C_2}{\sigma_X^2+\sigma_Y^2+C_2} \cdot \frac{\sigma_{X Y}+C_3}{\sigma_X \sigma_Y+C_3}
\end{equation}
where, $C_1, C_2, C_3$ are constants used to avoid system errors. $\mu$ and $\sigma$ represent the mean and variance of two image windows, respectively, while the overall score corresponds to the mean.

\subsection{Experimental Results}

We compared the performance of the proposed method on the RICE1 dataset with other GAN-based cloud removal approaches, as shown in Table \ref{tab:performance_rice-all}. The results of the reference methods are collected from \cite{pan2020cloud}. The proposed method demonstrates a clear improvement over other GAN-based approaches in terms of both PSNR and SSIM. The improvement in SSIM (0.976) highlights its ability to maintain structural integrity, while the high PSNR (33.659) reflects enhanced image quality. These results suggest that the proposed method is more effective at addressing the complexities of cloud removal in remote sensing images.

Apart from RICE1, we also evaluate the proposed method on the RICE2 dataset. and results are also shown in Table~\ref{tab:performance_rice-all}. Results of reference method are collected from \cite{pan2020cloud}. The proposed achieves the best results in both PSNR and SSIM, showcasing its superiority in generating cloud-free images with excellent quality and strong structural preservation.

\begin{table}
    \centering
    \caption{Performance comparison between the proposed method and other GAN-based techniques on the RICE1 and RICE2 datasets. }
    \begin{tabular}{p{2.5cm}p{1cm}p{1cm}|p{1cm}p{1cm}}
    \toprule
    &\multicolumn{2}{c|}{\textbf{RICE1}} &\multicolumn{2}{c}{\textbf{RICE2}}\\\toprule
        \textit{Model}            & \textit{PSNR} & \textit{SSIM} & \textit{PSNR} & \textit{SSIM} \\\toprule
        Conditional GAN  & 26.547 & 0.903 & 25.386 & 0.811\\\hline
        Cycle GAN        & 25.880 & 0.893& 23.910 & 0.793\\\hline
        SpA GAN          & 30.232 & 0.954& 28.368 & 0.906\\\toprule
        The Proposed Model & \textbf{33.659} & \textbf{0.976}& \textbf{34.056} & \textbf{0.955}\\\bottomrule
    \end{tabular} \vspace{-5pt}
    \label{tab:performance_rice-all}
\end{table}

    

Furthermore, we evaluated our proposed method against state-of-the-art cloud removal approaches using SSIM as the performance metric for two benchmark datasets, as presented in Table \ref{tab:SoTA}. However, it is important to note that the authors \cite{10466744} of the compared method -- \textit{CMNet} -- and remaining methods did not disclose critical details about their dataset configurations, such as the specific splits used for training, testing, and validation. This lack of transparency makes a direct, fair comparison challenging. Despite these limitations, the SSIM values indicate that our proposed method achieves competitive performance, demonstrating its effectiveness in addressing the cloud removal task even under these less-than-ideal comparative conditions.

\begin{table}
    \centering
    \caption{SSIM of state-of-the-art approaches.}
    \setlength{\tabcolsep}{1.5pt} 
    \begin{tabular}{ccccccc|c}
    \toprule
        \textbf{Method} & \textbf{DCP} & \textbf{YUV-GAN} & \textbf{MS-GAN} & \textbf{RCAN} & \textbf{CVAE} & \textbf{CMNet} & \textbf{Ours}\\\toprule
        RICE1 & 0.769 & 0.888 & 0.932 & 0.935 & 0.886 & 0.970 & 0.976\\\hline
        RICE2 & 0.645 & 0.819 & 0.881 & 0.840 & 0.919 & 0.926 & 0.955\\\bottomrule
    \end{tabular}
    
    \label{tab:SoTA}
\end{table}

\begin{figure}
    \centering
    \includegraphics[width=\linewidth]{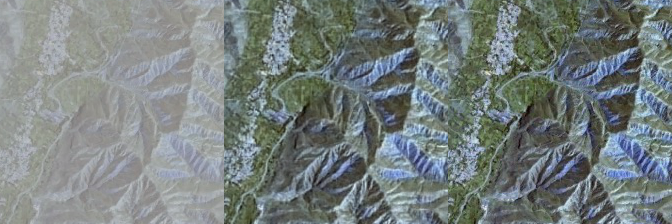}

    \vspace{5pt}
    \includegraphics[width=\linewidth]{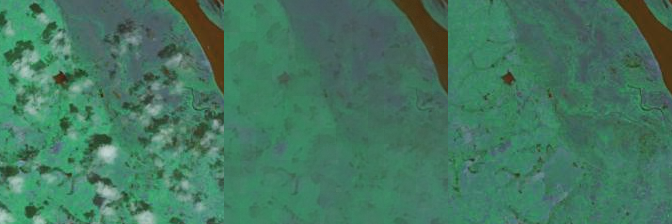}
    \caption{Visual results on the RICE1 (top) and RICE2 (bottom) datasets. From left to right are the cloud-covered image, the generated cloud-free image, and the ground truth cloud-free image.}
    \label{fig:visual_result}\vspace{-0.25cm}
\end{figure}

Some visual examples are displayed in Figure \ref{fig:visual_result}. The generated cloud-free images efficiently remove clouds and closely resemble the ground truth cloud-free images in both the RICE1 and RICE2 datasets.

\section{Conclusion}

In this study, we introduced a novel deep transfer learning approach that adapts a model pre-trained on a large natural image dataset for image reconstruction to the task of cloud removal in remote sensing. The proposed method leverages a GAN framework with a patch-wise discriminator, which simplifies the discriminator's task by focusing on localized regions of the image. This design enhances the discriminator's accuracy and, consequently, the overall performance of the model.  

Our approach significantly outperforms existing GAN-based methods for cloud removal and demonstrates competitive results compared to state-of-the-art techniques, as evaluated using SSIM. This highlights the effectiveness of integrating deep transfer learning and patch-based evaluation for addressing the complexities of remote sensing imagery.  

In future work, we aim to investigate the potential of larger vision and language models for remote sensing tasks. Given their remarkable performance across various domains, these models hold promise for advancing cloud removal and other remote sensing applications.

\newpage\small
\bibliographystyle{IEEEtranN}
\bibliography{references}

\end{document}